\documentclass[letterpaper, 10 pt, conference]{ieeeconf}
\IEEEoverridecommandlockouts                      
\usepackage{afterpage}
\usepackage{amsmath,amsfonts}
\usepackage{algorithmic}
\usepackage{algorithm}
\usepackage{array}
\usepackage{graphicx}
\usepackage[caption=false, font=small, labelfont=sf,textfont=sf]{subfig}
\usepackage{textcomp}
\usepackage{stfloats}
\usepackage{url}
\usepackage{verbatim}
\usepackage{graphicx}
\usepackage{cite}
\usepackage{xcolor}
\usepackage{soul}
\usepackage{multirow}
\usepackage{siunitx}
\usepackage{soul}
\usepackage{fancyhdr}
\usepackage{hyperref}
\usepackage{booktabs}

\begin{document}

\title{OmniUnet: A Multimodal Network for Unstructured Terrain Segmentation on Planetary Rovers Using RGB, Depth, and Thermal Imagery}

\author{R.~Castilla-Arquillo$^{1,2}$, C.~J.~Pérez-del-Pulgar$^{2}$, L.~Gerdes$^{2}$, A.~Garcia-Cerezo$^{2}$, and M. Olivares-Mendez$^{1}$
\thanks{$^{1}$R. Castilla-Arquillo and M. Olivares-Mendez are with the Space Robotics (SpaceR) Research Group, SnT, University of Luxembourg, Luxembourg, Luxembourg (e-mail:raul.castilla@uni.lu).}
\thanks{$^{2}$R. Castilla-Arquillo, C. J. Pérez-del-Pulgar, L.~Gerdes, and A. García Cerezo are with the Department of Automation and Systems Engineering, Universidad de Málaga, Andalucía Tech, 29070 Málaga, Spain (e-mail:carlosperez@uma.es).}}

\maketitle
\begin{abstract}

Robot navigation in unstructured environments requires multimodal perception systems that can support safe navigation. Multimodality enables the integration of complementary information collected by different sensors. However, this information must be processed by machine learning algorithms specifically designed to leverage heterogeneous data. Furthermore, it is necessary to identify which sensor modalities are most informative for navigation in the target environment. In Martian exploration, thermal imagery has proven valuable for assessing terrain safety due to differences in thermal behaviour between soil types. This work presents OmniUnet, a transformer-based neural network architecture for semantic segmentation using RGB, depth, and thermal (RGB-D-T) imagery. A custom multimodal sensor housing was developed using 3D printing and mounted on the Martian Rover Testbed for Autonomy (MaRTA) to collect a multimodal dataset in the Bardenas semi-desert in northern Spain. This location serves as a representative environment of the Martian surface, featuring terrain types such as sand, bedrock, and compact soil. A subset of this dataset was manually labeled to support supervised training of the network. The model was evaluated both quantitatively and qualitatively, achieving a pixel accuracy of 80.37\,\% and demonstrating strong performance in segmenting complex unstructured terrain. Inference tests yielded an average prediction time of 673\,ms on a resource-constrained computer (Jetson Orin Nano), confirming its suitability for on-robot deployment. The software implementation of the network and the labeled dataset have been made publicly available to support future research in multimodal terrain perception for planetary robotics.
\end{abstract}

\section{INTRODUCTION} 
\label{sec:introduction}

Robot navigation in unstructured environments presents significant challenges, such as steep slopes, uneven surfaces, and loose or rocky terrain~\cite{Wijayathunga2023}. These challenges are especially relevant in planetary exploration, where limited communication and the absence of physical access mean that a single failure can impact the progress of a scientific mission~\cite{gonzalez2018slippage}. To address these challenges, autonomous systems must rely on perception systems that can remotely assess their surroundings and enable safe navigation. A core function of such perception systems is terrain segmentation, which divides the landscape into regions with similar surface characteristics to distinguish traversable areas, detect obstacles, and identify potential scientific targets.

\begin{figure}
    \centering
    \includegraphics[width=0.35\textwidth]{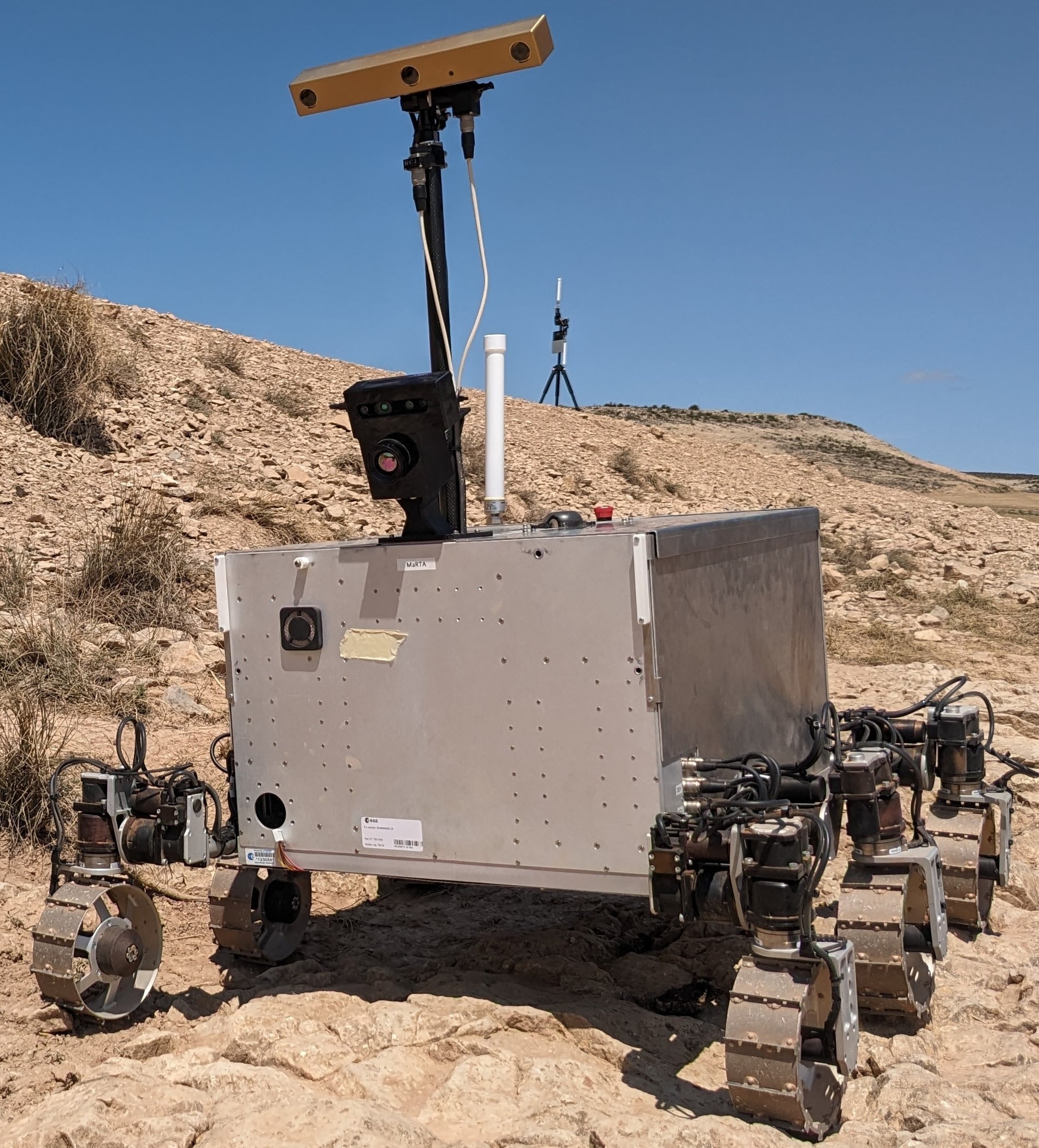}
    \caption{\small MaRTA rover in the Bardenas semi-desert, equipped with the custom multimodal sensor unit.}
    \label{fig:marta-rover}
\end{figure}

In unstructured scenarios, perception systems often incorporate sensors of different modalities. Multimodal sensor fusion enables systems to overcome the limitations of individual sensors by combining their complementary strengths~\cite{zhang2023perception}. A notable example of a multimodal sensor suite is the ScienceCam, integrated into the Light Weight Rover Unit (LRU)\cite{Wedler2017}, which includes grayscale, RGB, hyperspectral, and thermal cameras for terrain segmentation. Similarly, Mars rovers such as Curiosity~\cite {grotzinger2012mars} and Perseverance~\cite{Bell2021} utilize multimodal sensors to support both navigation and scientific research.

Among the sensor modalities used in perception systems, thermal imaging offers some advantages over conventional approaches, such as RGB or stereo vision. It is particularly effective in conditions with low lighting or strong glare, where it can enhance the reliability of perception. In unstructured environments, thermal surface properties such as thermal inertia provide valuable information for estimating terrain traversability. Under solar heating, granular sandy soils exhibit higher surface temperatures than less granular compacted soils~\cite{gonzalez2017}. This relationship between thermal characteristics and traversability is especially relevant on planets like Mars, where the low atmospheric pressure increases thermal contrasts and supports the estimation of rover slippage~\cite{cunningham2019improving, castilla2023thermal}.

Achieving accurate terrain segmentation from multimodal inputs requires machine learning models that can effectively process and integrate heterogeneous visual data. Transformer-based architectures with shifted windows (Swin)~\cite{liu2021swin} and multi-scale attention mechanisms~\cite{fan2021multiscale} have recently gained attention for their ability to handle such complexity. A notable example is Omnivore~\cite{girdhar2022omnivore}, a Swin-based architecture that supports training across different input types without requiring simultaneous availability. Additionally, models that combine Swin backbones with U-Net-style decoders~\cite{ronneberger2015u}, such as Swin-Unet~\cite{cao2022swin} and UNetFormer~\cite{wang2022unetformer}, have shown strong performance in single-modality segmentation tasks, including applications in medical and urban environments. However, these architectures are not specifically designed for multimodal integration and often require adaptation and optimization to run efficiently on embedded platforms used in mobile robotic systems.

\begin{figure*}[]
    \centering
    \includegraphics[width=1\textwidth]{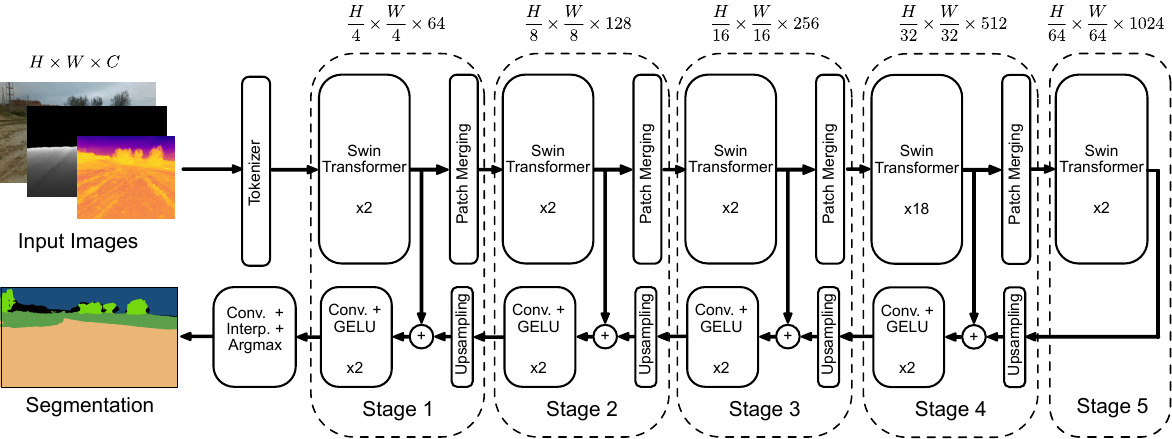}
    \caption{\small Diagram of the OmniUnet network.}
    \label{fig:omniunet-diagram}
\end{figure*}

Several multimodal datasets have been developed to support research in unstructured navigation. The RUGD dataset~\cite{wigness2019rugd} comprises video sequences captured by a mobile robot equipped with a LiDAR sensor and a monocular camera, covering a diverse range of terrain classes. RELLIS-3D~\cite{jiang2021rellis} provides synchronized LiDAR and stereo camera data collected from a mobile platform in diverse outdoor environments, with semantic annotations for various terrain types. The CAVS dataset~\cite{sharma2022cat} provides paired image and LiDAR data recorded under varying lighting conditions, vegetation densities, and atmospheric states. Finally, the BASEPROD \cite{gerdes2024baseprod} dataset contains RGB, depth, and thermal data collected during rover traverses in the Bardenas semi-desert in northern Spain, a location considered a representative environment of the Martian surface. While it is well-suited for evaluating multimodal perception models and supports the sensor configuration used in this study, it does not include segmentation masks, limiting its use for supervised training tasks.

This work presents OmniUnet, a novel transformer-based neural network architecture designed to generate segmentation images of unstructured environments using RGB, depth, and thermal (RGB-D-T) imagery. To support this approach, we developed a custom multimodal sensor housing to mount stereo and thermal cameras on the Martian Rover Testbed for Autonomy (MaRTA), a half-scale model of the six-wheeled ExoMars Rosalind Franklin rover, shown in Fig.~\ref{fig:marta-rover}. This sensor configuration was used to record a new multimodal dataset in the Bardenas semi-desert, capturing a range of terrain types including bedrock, sand, and compact soil \cite{gerdes2024baseprod}. As part of this work, we provide labeled segmentation masks of the dataset\cite{Castilla2025SegmentationData}, which are publicly available on Zenodo\footnote{\url{https://doi.org/10.5281/zenodo.15496884}} to support further research in multimodal terrain perception. Additionally, the software for training and deploying the OmniUnet model is available on GitHub\footnote{\url{https://github.com/spaceuma/OmniUnet}}, enabling reproducibility and facilitating its use in related robotic perception tasks.

\section{MULTIMODAL TERRAIN SEGMENTATION}

OmniUnet is a transformer-based neural network architecture that combines the Omnivore backbone\cite{girdhar2022omnivore} for feature extraction with the U-Net decoding strategy~\cite{ronneberger2015u} to produce segmentation images of unstructured environments. The architecture is designed to exploit the complementary characteristics of multiple sensing modalities, generating terrain segmentation maps suitable for autonomous rover navigation. To support effective learning from heterogeneous input data, a dedicated training strategy was also developed.

\begin{figure}
    \centering
    \subfloat[\label{sfig:multi-housing-render}]{\includegraphics[height=0.22\textwidth]{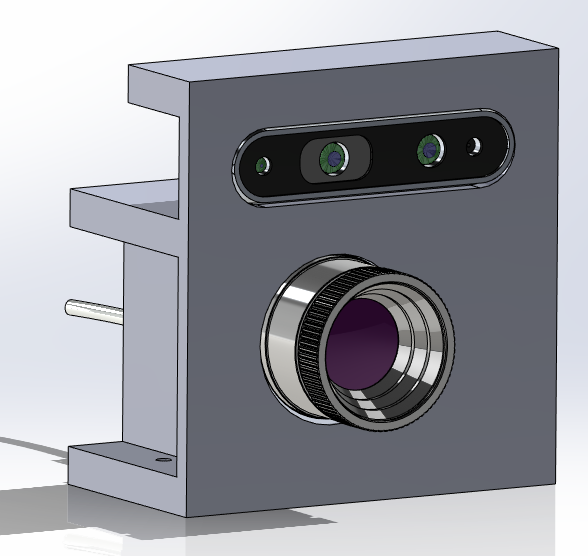}}
    \hspace{0.1cm}
    \subfloat[\label{sfig:multi-housing-printed}]{\includegraphics[height=0.22\textwidth]{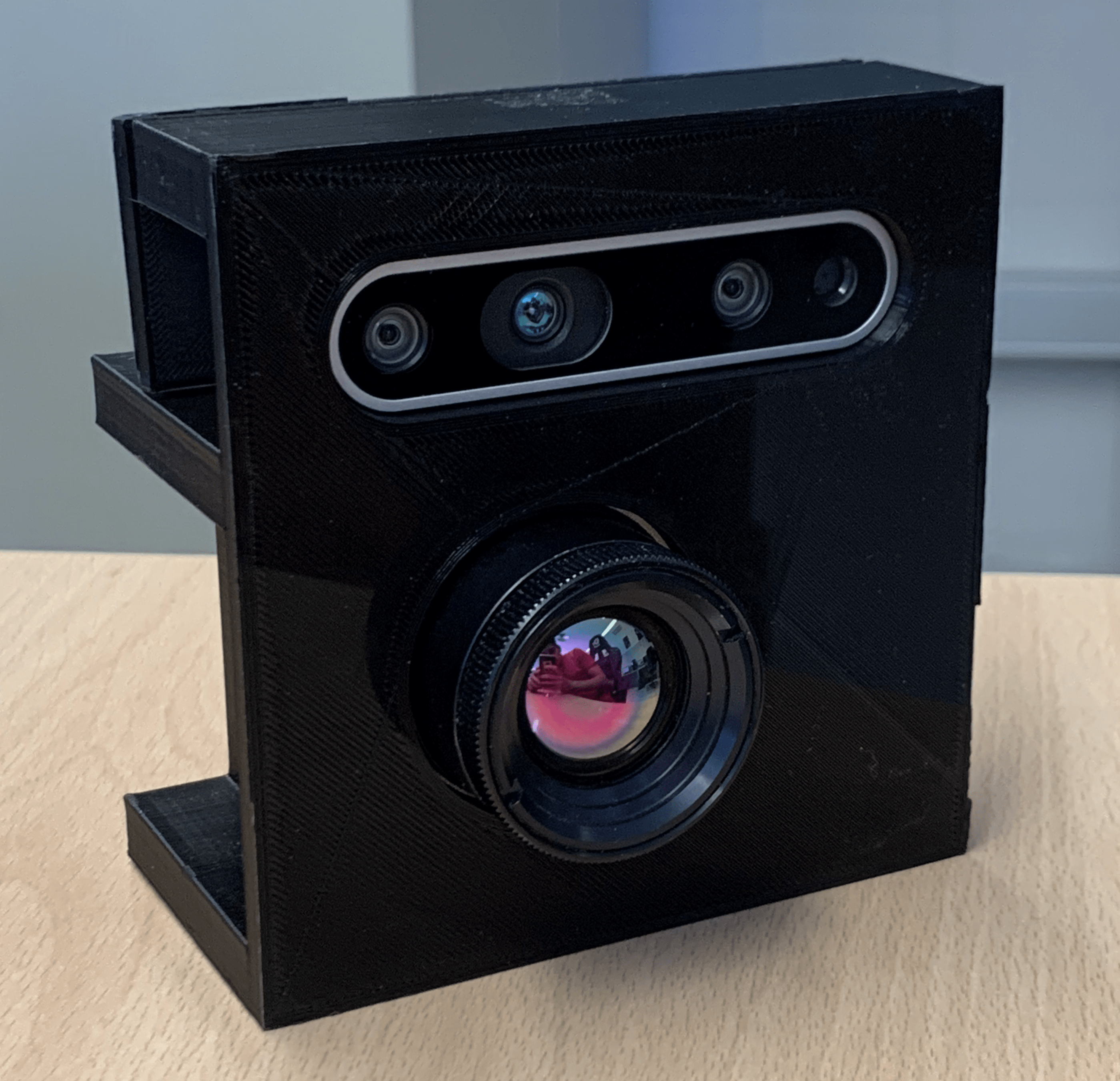}}
    \caption{\small Housing for the multimodal sensors composed of the Realsense (top side) and Optris (bottom side) cameras: a) Rendered 3D model design; b) 3D printed model.}
    \label{ffig:multi-housing}
\end{figure}

\subsection{Architecture}
OmniUnet accepts five-channel input images that combine RGB, depth, and thermal modalities. The architecture, shown in Fig.~\ref{fig:omniunet-diagram}, is designed to be flexible and can be extended to accommodate additional input modalities if needed. While training requires high-performance GPU resources due to the model's computational complexity, inference can be performed in real time on embedded GPUs, making it suitable for deployment in field robotics applications.

The multimodal image processing begins with a tokenizer layer that converts the input images into patch embeddings, which are then passed to the network's transformer-based backbone. This step facilitates the integration of information from the different modalities, allowing the network to extract relevant features and correlations across input types. Using a unified architecture for all modalities eliminates the need for separate models, supporting a more scalable and maintainable design.

The patch embeddings provided by the tokenizer are processed through a sequence of transformer blocks that extract hierarchical features while maintaining the number of tokens. The Swin transformer employs a shifted window attention mechanism, where the input is divided into small, non-overlapping windows, and self-attention is applied within each window. To allow interaction between windows, the window positions are shifted in alternating layers so that adjacent regions overlap, enabling neighboring windows to contribute to the attention computation. This method preserves local context while allowing global information flow, making it suitable for processing diverse data types and capturing both spatial and temporal relationships among them.

In the decoding stage of OmniUnet, each feature map is combined with the upsampled output from the stage above. The fused representation is then processed through two consecutive $3 \times 3$ convolutional layers followed by Gaussian Error Linear Unit (GELU) activation before being passed to a lower decoding stage. Finally, the output of Stage 1 produces the network's logits, which are converted into a probability distribution over all the target classes, resulting in the predicted segmentation mask of the rover's surroundings.

\subsection{Training strategy}

During training, the Sørensen–Dice coefficient, also known as the Dice coefficient, is used to assess the similarity between predicted and ground-truth segmentation masks, providing a balanced measure of precision and recall. For each class $c$, the Dice coefficient $D_{c}$ is computed as:

\begin{equation}
D_{c} = \frac{2 | P_{c} \cap G_{c} |}{| P_{c} | + | G_{c} |} = \frac{2 TP_{c}}{2 TP_{c} + FP_{c} + FN_{c}},
\end{equation}

where $P_{c}$ and $G_{c}$ represent the predicted and ground truth segmentation masks for class $c$, respectively, and $TP_{c}$, $FP_{c}$, and $FN_{c}$ are the corresponding true positive, false positive, and false negative pixel counts.

The loss function, $\mathcal{L}$, aimed at minimization during training, combines the cross-entropy loss and the Dice loss. It produces decimal values ranging from 2 to 0, where a value of 0 indicates a perfect match between the predicted image and the ground truth:

\begin{equation}
\mathcal{L} = -\sum_{c=0}^N G_{c}\log(P_{c}) + \sum_{c=1}^N (1 - D_{c}),
\end{equation}

where the first term calculates the total cross-entropy loss between the predicted masks and the target masks across all classes ($N$), while the second term sums up the Dice loss for all classes, excluding the background or void class.

\begin{table*}
\centering
\setlength{\tabcolsep}{3pt} 
\renewcommand{\arraystretch}{1.5} 
\centering
\vspace{3mm}
\small
\caption{Characteristics and metrics of the OmniUnet network evaluated on the RUGD dataset.}
\begin{tabular}{l m{2mm} ccc m{2.5mm}  m{10mm} m{10mm} m{10mm}}
& &  \multicolumn{3}{c}{Dataset Images} &  & \multicolumn{3}{c}{Average Metrics} \\ \cmidrule(lr){3-6} \cmidrule(lr){7-9}
 Classes & & Training & Validation & Total & & Total PA & Mean PA & Mean \newline IoU \\
\hline
 25 & & 5948 & 1487 & 7435 & & \textbf{92.58} & 39.72 & 34.90 \\
\hline
\end{tabular}
\label{tab:dataset-omniunet-rugd}
\end{table*}

\vspace{5 mm}

\begin{figure*}
\centering
\captionsetup{width = 1\linewidth}
\includegraphics[width=0.75\textwidth]{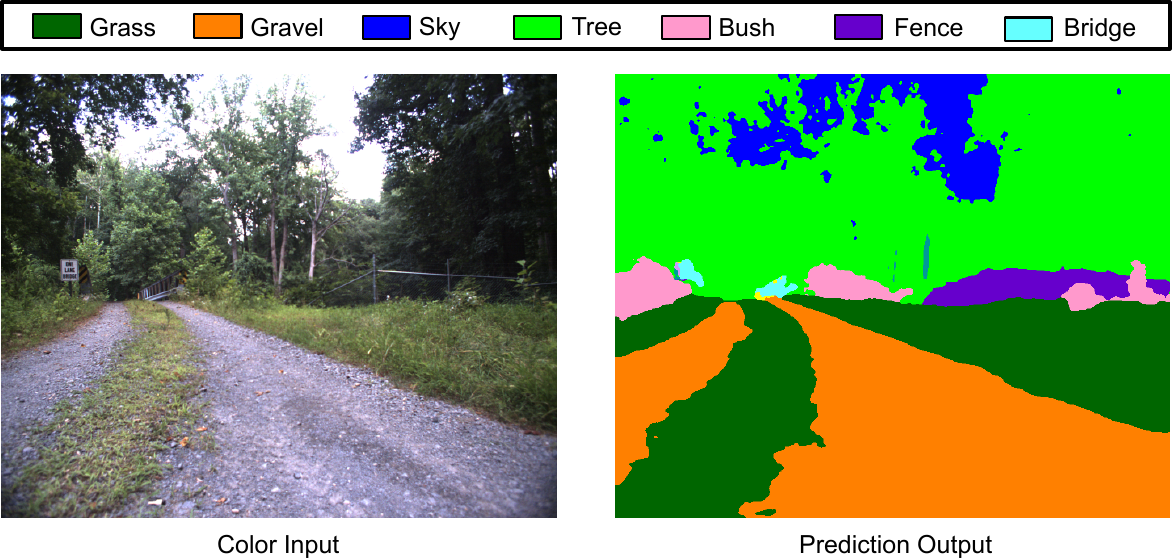}
\caption{\small Color image from the RUGD dataset segmented by OmniUnet.}
\label{ffig:rugd_example}
\end{figure*}

\section{IMPLEMENTATION AND RESULTS}

A  custom multimodal sensor housing was developed to mount stereo and thermal cameras on MaRTA. This sensor configuration was used to record the BASEPROD multimodal dataset \cite{gerdes2024baseprod} in the Bardenas semi-desert in northern Spain. OmniUnet was validated using the BASEPROD dataset with RGB-D-T imagery, and its performance was assessed using both quantitative metrics and qualitative analysis. To demonstrate the model's flexibility, OmniUnet was also trained and tested using only RGB images from the RUGD dataset, confirming its capability to operate across various input configurations and sensor modalities.

\subsection{Multimodal sensor}

A custom multimodal sensor housing was developed using 3D printing to integrate a Realsense D435i stereo camera and an Optris PI-640i thermal camera, as shown in Fig.~\ref{ffig:multi-housing}. The housing was designed to ensure proper alignment between the depth plane of the Realsense and the optical axis of the thermal sensor, facilitating accurate multimodal image registration. However, due to the different fields of view (FOV) of the two cameras, the alignment process results in small black areas along the left and right borders of the fused images. The Realsense camera includes an RGB sensor capable of capturing images at a resolution of up to \numproduct{1920 x 1080} pixels with a FOV of \qtyproduct{69 x 42}{\degree}, and provides stereoscopic depth images at \numproduct{1280 x 720} resolution with a FOV of \qtyproduct{87x58}{\degree} and depth accuracy of less than \qty{2}{\%} at a distance of 2 meters. The Optris PI-640i thermal camera operates in the Long-Wave Infrared (LWIR) spectral range from \qty{8}{\um} to \qty{14}{\um}, using uncooled microbolometer technology to produce thermal images at $640\times480$ resolution. It measures temperatures from \qtyrange{-20}{900}{\celsius} with a thermal sensitivity of \qty{0.04}{\celsius} and is equipped with a germanium lens offering a \qtyproduct{60x45}{\degree} field of view.

\subsection{Training and evaluation}

For validation, both the RUGD and BASEPROD datasets were employed. As there is currently no publicly available labeled multimodal dataset of unstructured environments comparable to ours, the network was initially trained and evaluated using the RUGD dataset to establish a performance baseline. This dataset contains approximately 7,500 labeled RGB images of unstructured outdoor scenes annotated across 25 semantic classes. The Bardenas dataset, specifically labeled for this study, includes data from 24 rover traverses, ranging in length from \qty{6.85}{\m} to \qty{202}{\m}, covering a total of approximately \SI{1.7}{\km}. A total of 36,000 multimodal images were recorded, of which 950 representative samples were manually labeled for training and 190 were set aside for validation. This dataset comprises eight terrain classes: void, compact, bedrock, sandy, gravel, rock, bush, and grass.

Training was conducted on an NVIDIA DGX Station equipped with a Tesla V100 DGX GPU, featuring \SI{32}{\giga\byte} of VRAM. The network was trained for $50$ epochs with a batch size of $16$ and a learning rate of $2 \times 10^{-5}$. The optimization was performed using a composite loss function combining Dice loss and cross-entropy loss. For each session, the model weights corresponding to the epoch with the lowest validation loss were retained. All datasets were divided into training and validation sets using an 80/20 split.

To simulate realistic on-board deployment, the network was evaluated on an NVIDIA GeForce RTX 2070 with \qty{8}{\giga\byte} of VRAM, achieving an average inference time of \qty{137.50}{\ms}. Additionally, OmniUnet was tested on a Jetson Orin Nano featuring an ARM processor running at \qty{1.5}{\giga\hertz}, \qty{8}{\giga\byte} of RAM, and an AI performance of 40 TOPS. This device was selected for its low power consumption (\qty{15}{\watt}) and suitability for resource-constrained robotic platforms. On this hardware, the network achieved an average inference time of \qty{673}{\ms} per multimodal image. The following metrics were used to measure the performance of the network:

\begin{itemize}
\item Pixel Accuracy (PA): This metric measures the proportion of correctly classified pixels by comparing the number of accurate predictions to the total number of pixels for a given class. It also accounts for true negative pixels, denoted as $TN_{c}$. The final pixel accuracy can be reported in two ways: total PA, which considers all correctly classified pixels regardless of their class, and mean PA, which is the average of the pixel accuracy metrics calculated for each class. It is defined as follows:

\begin{equation}
PA = \frac{TP_{c} + TN_{c}}{TP_{c} + TN_{c} + FP_{c} + FN_{c}} ,
\end{equation}

\item Intersection over Union (IoU): Also known as the Jaccard index, this metric measures the overlap between the predicted and ground-truth regions. It is calculated as:

\begin{equation}
IoU = \frac{TP_{c}}{TP_{c} + FP_{c} + FN_{c}} ,
\end{equation}

\end{itemize}

\begin{table*}
\centering
\setlength{\tabcolsep}{3pt} 
\renewcommand{\arraystretch}{1.5} 
\centering
\vspace{3mm}
\caption{Metrics of the OmniUnet network evaluated on the Bardenas dataset.}
\small
\begin{tabular}{cccccccc m{2mm} m{10mm} m{10mm} m{10mm}}
  \multicolumn{8}{c}{Class Pixel Accuracy} &  & \multicolumn{3}{c}{Average Metrics} \\ \cmidrule(lr){1-8} \cmidrule(lr){10-12}
 Void & Compact & Grass & Bedrock & Sandy & Gravel & Rock & Bush & & Total PA & Mean PA & Mean \newline IoU \\
\hline
90.48  & 77.44  & 39.04 & 28.30 & 26.57 & 51.83 & 18.40 & 40.68 & & \textbf{80.37} & 46.59 & 38.77 \\
\hline
\end{tabular}
\label{tab:baseprod-omniunet-network}
\end{table*}

\begin{figure*}
\centering
\includegraphics[width= 1\textwidth]{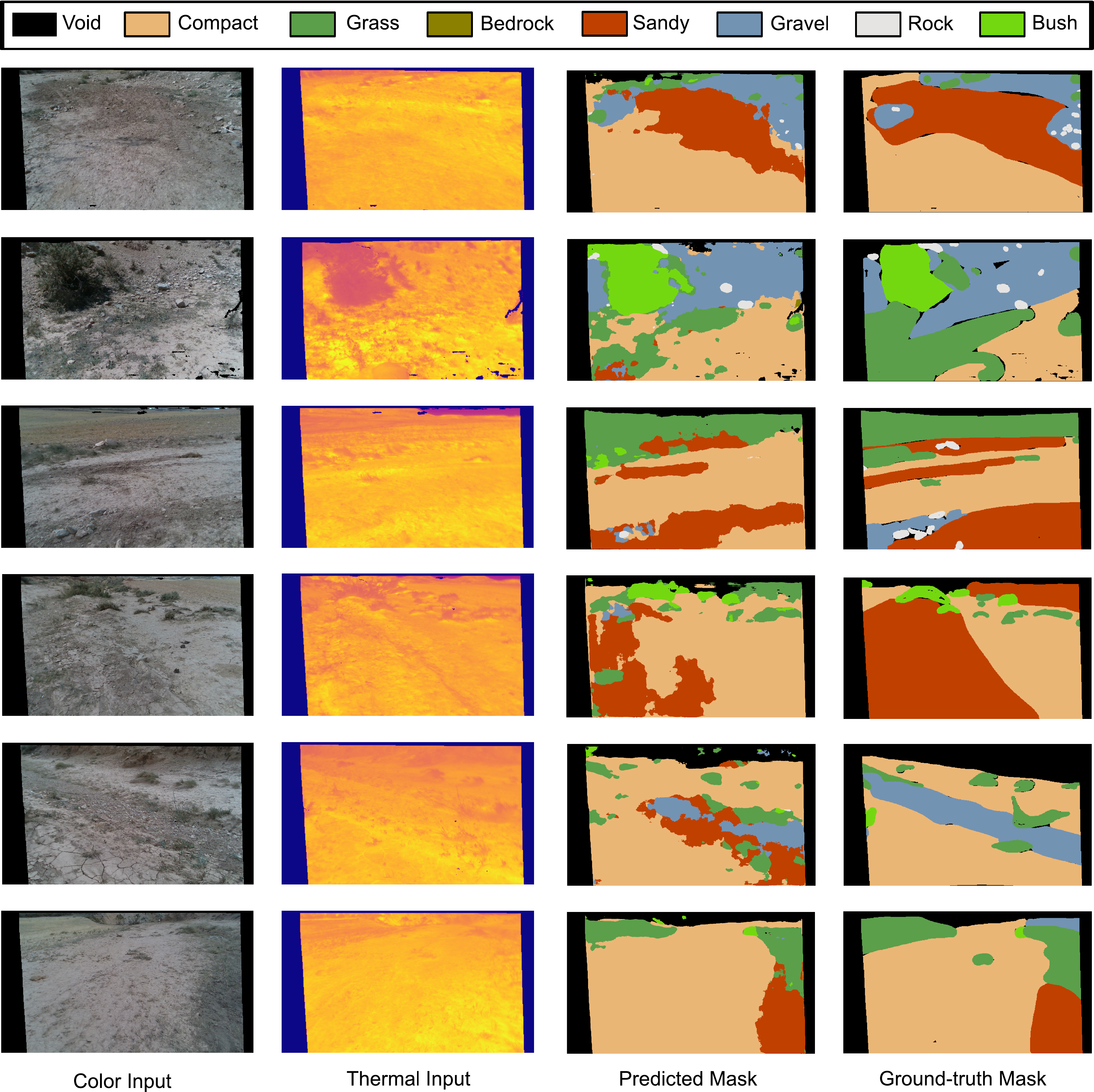}
\caption{\small Segmentation masks produced by OmniUnet using the Bardenas dataset alongside their corresponding inputs.}
\label{fig:omniunet-baseprod}
\end{figure*}

\subsection{Results}

The results obtained by OmniUnet on the RUGD validation set are presented in Table~\ref{tab:dataset-omniunet-rugd}, where its characteristics and average performance metrics are provided. An example of a segmented image from this dataset is shown in Figure~\ref{ffig:rugd_example}. For the labeled BASEPROD dataset, Table~\ref{tab:baseprod-omniunet-network} reports per-class pixel accuracy and average metrics, excluding the void class to reflect more realistic performance. Figure~\ref{fig:omniunet-baseprod} illustrates the segmentation masks produced by the network, alongside the corresponding color and thermal inputs.

In terms of quantitative results, OmniUnet achieved a pixel accuracy of \qty{92.58}{\%} on the RUGD dataset and \qty{80.37}{\%} on the BASEPROD multimodal dataset. In BASEPROD, the model demonstrated strong performance in identifying surfaces relevant to safe navigation. It detected compact terrain, considered the safest to traverse, with an accuracy of \qty{77.44}{\percent}, and achieved \qty{51.83}{\%} accuracy in identifying gravel terrain. Nonetheless, drawing general conclusions based on per-class accuracy remains challenging due to the complexity of unstructured terrain and the presence of transition zones between terrain types. For obstacle detection, the model identified bushes with an accuracy of \qty{40.68}{\%}, while rock detection proved more difficult, achieving \qty{18.40}{\%} accuracy, likely due to the variability in the thermal signatures of rocks. Qualitative results further support the model's effectiveness, highlighting its ability to distinguish between sandy and compact terrain, which differ in thermal behavior. Sandy areas generally exhibit higher surface temperatures due to their lower thermal inertia, allowing the model to exploit this contrast in the thermal domain for more accurate segmentation. In some cases, the network's output even outperformed the ground truth annotations in identifying transition areas between sandy and compact terrain.

\section{CONCLUSIONS AND FUTURE WORK}
\label{sec:conclusions}

This work presented OmniUnet, a transformer-based neural network architecture developed to generate semantic segmentation images of unstructured environments using RGB, depth, and thermal (RGB-D-T) imagery. To support this approach, a custom multimodal sensor housing was designed and integrated into the MaRTA rover, enabling the collection of a labeled RGB-D-T dataset in the Bardenas semi-desert. The network demonstrated strong capabilities in using multimodal imagery for segmenting unstructured terrains, achieving a pixel accuracy of \qty{80.37}{\%} across diverse terrain types. This capability is expected to be even more effective in Martian environments, where thermal inertia contrasts between terrain types are more pronounced. Furthermore, OmniUnet was successfully deployed on a resource-constrained platform (Jetson Orin Nano), achieving an inference time of \SI{673}{\ms} per multimodal image.

Future work will focus on enhancing the accuracy and robustness of segmentation in complex natural environments. One of the main challenges lies in managing a growing number of terrain classes, particularly when distinguishing visually similar surfaces and obstacles. To address this, future developments may include the integration of a dedicated neural network module specialized in obstacle detection. Additional efforts will focus on designing improved multimodal sensor housings, specifically optimized for stereo-thermal imaging in low-light conditions, with the aim of further enhancing perception reliability in both terrestrial and planetary applications.

\section*{ACKNOWLEDGMENTS}
This work was partially supported by the European Space Agency under activity no.\
4000140043/22/NL/GLC/ces, and by the Spanish national government grants no.\ PID2021-122944OB-I00, titled SAR4.0 – Leapfrogging to a New Paradigm in Cooperative Human-Robot Cyber-Physical Systems for Search and Rescue, and no.\ PID2024-160373OB-C21, titled Hybrid Perception and Navigation for Planetary Exploration.

\bibliographystyle{IEEEtran}
\bibliography{references}

\vfill
\end{document}